# CQ-VAE: Coordinate Quantized VAE for Uncertainty Estimation with Application to Disk Shape Analysis from Lumbar Spine MRI Images

Linchen Qian
*Department of Computer Science*
*University of Miami*
Coral Gables, FL, USA
lxq93@miami.edu

Jiasong Chen
*Department of Computer Science*
*University of Miami*
Coral Gables, FL, USA
jxc2206@miami.edu

Timur Urakov
*Department of Neurological Surgery*
*University of Miami*
Coral Gables, FL, USA
turakov@med.miami.edu

Weiyong Gu
*Department of Mechanical and Aerospace Engineering*
*University of Miami*
Coral Gables, FL, USA
wgu@miami.edu

Liang Liang
*Department of Computer Science*
*University of Miami*
Coral Gables, FL, USA
liang@cs.miami.edu

*Abstract*—Ambiguity is inevitable in medical images, which often results in different image interpretations (e.g. object boundaries or segmentation maps) from different human experts. Thus, a model that learns the ambiguity and outputs a probability distribution of the target, would be valuable for medical applications to assess the uncertainty of diagnosis. In this paper, we propose a powerful generative model to learn a representation of ambiguity and to generate probabilistic outputs. Our model, named Coordinate Quantization Variational Autoencoder (CQ-VAE) employs a discrete latent space with an internal discrete probability distribution by quantizing the coordinates of a continuous latent space. As a result, the output distribution from CQ-VAE is discrete. During training, Gumbel-Softmax sampling is used to enable backpropagation through the discrete latent space. A matching algorithm is used to establish the correspondence between model-generated samples and "ground-truth" samples, which makes a trade-off between the ability to generate new samples and the ability to represent training samples. Besides these probabilistic components to generate possible outputs, our model has a deterministic path to output the best estimation. We demonstrated our method on a lumbar disk image dataset, and the results show that our CQ-VAE can learn lumbar disk shape variation and uncertainty.

## I. Introduction

The uncertainty of object geometry given the image of an object (e.g., a lumbar disk) is related to model uncertainty (a.k.a. epistemic uncertainty) and data uncertainty (a.k.a. aleatory uncertainty). Traditional methods focus on prediction intervals [1], which are unable to depict uncertainty of high dimensional outputs, such as the shape of an object. Recently, Bayesian framework was employed by Garg et al. [2] and Jena et al. [3] for uncertainty estimation in image segmentation task. A VAE-based U-Net was proposed by Kohl et al. [4] to generate multiple possible mask hypotheses to describe segmentation uncertainty. This VAE-based U-Net samples a random feature map from a generative model and concatenates this map to the last activation map of U-Net to generate a possible segmentation mask. Due to its unique structure, it cannot be used for shape regression tasks.

For medical image analysis, if the object of interest (e.g. human organs) has a well-defined shape, shape regression is preferred than image segmentation which may output fragmented regions and irregular boundaries. Here, shape regression refers to predicting an object's shape, represented by a 2D/3D contour/mesh, from the object's images. Let's use the diagnosis of lumbar disk degeneration from MRI images as a concrete example [5]. Firstly, a human expert delineates the boundary of each lumbar disk in the MRI images using some image annotation software (e.g. MATLAB). Then, the clinical features of these disks will be extracted from shape and image intensity, which can be easily done by computer algorithms. Finally, Doctors will make a diagnosis of the lumbar spine. However, due to ambiguity in medical images, different human experts may have different opinions about a disk's shape, and the usual solution could be taking the average of the shapes obtained by different human experts or choosing the most confident one. These "solutions" do not reduce the uncertainty but merely ignore it. Obviously, shape uncertainty leads to diagnosis uncertainty: if the uncertainty is small, then it may be ignored; and if it is large, then the patient may need to take another MRI imaging with a higher resolution and contrast dye. Thus, it would be very useful to quantify the uncertainty, preferably by automated algorithms. We note that although an end-to-end machine learning solution could be made possible (e.g. input image and output disk degeneration level), the doctors would prefer the disk features calculated from shapes, which are explainable and intuitive.

A probability distribution of shapes is the best way to quantify the uncertainty of shape in images. From the shape probability distribution, a large number of possible shapes can be generated and analyzed, and then the distribution (i.e. uncertainty) of the features derived from the shapes can be quantified. To design a deep neural network (DNN) for uncertainty quantification, we need to decide whether to use a discrete or continuous distribution. Gaussian model or Gaussian mixture model (GMM) is a natural choice to represent a continuous distribution. Currently, VAE [6] can be used to learn a Gaussian distribution in the latent space, but it has a severe limitation: the covariance of the Gaussian distribution is an identity matrix multiplied by a scalar which is a function of the input. A straightforward extension is to use a GMM with non-diagonal covariance matrices in the latent space. However, convergence problems arise when learning a GMM even in a low dimensional space by EM algorithm, which casts doubts about learning a GMM in a high dimensional space by DNNs. It has been shown that VAE covariance matrix being diagonal is the key for VAE to learn disentangled representation in the latent space [7]. Therefore, we decide to design a DNN to learn a discrete probability distribution of the output target (i.e. shape) using a discrete representation in the latent space.

There exist several VAEs with discrete representation in latent space. The first one [8] has binary and continuous latent variables and assumes a factorial Bernoulli distribution for the binary variables. To generate new samples, it needs to run a Markov chain over RBM and perform Gibbs sampling. This VAE is not a good candidate for learning discrete probability distribution of the output target because (1) Bernoulli distribution is too restrictive and it is not clear how to connect this latent distribution to the output distribution, and (2) Gibbs sampling is time-consuming. The second one [9] is called VQ-VAE, and it has a fixed number of code-vectors. VQ-VAE first transforms the input into feature vectors in a latent space and then performs vector quantization by replacing each feature vector of the input with its nearest neighbor in the set of code-vectors. VQ-VAE is also not a good candidate for our application because the posterior "probability" distribution of the latent variable $z$ given the input $x$, i.e. $p(z|x)$, is deterministic and therefore the (e.g. shape) uncertainty in $x$ (i.e. image) disappears in the latent space.

In this study, we propose a novel generative VAE that learns a discrete representation in the latent space and outputs a discrete probability distribution of the target. Since it performs coordinate quantization (CQ) in the latent space, we name it CQ-VAE. We demonstrate CQ-VAE on the application of lumbar disk shape analysis from MR images. As a by-product of this study, we developed a discrete autoencoder, named CQ-AE, and tested it on Fashion-MNIST dataset, which will help to explain CQ-VAE.

## II. METHODOLOGY

### A. Uncertainty

Let $x$ represent an image of an object and $y$ be the shape of the object. $\mathcal{P}(s|x)$ is the probability distribution (i.e. uncertainty) of the shape $s$ given the image $x$. Multiple models could be used to predict the shape, and each model is denoted by $\theta_j$. Assuming the models follow a distribution $\mathcal{P}(\theta)$ in a discrete model space $\Theta$ (e.g. DNN architecture space is often discretized for neural architecture search), then we obtain:

$$\mathcal{P}(s|x) = \sum_{\theta_j \in \Theta} \mathcal{P}(\theta_j)\mathcal{P}(s|\theta_j, x) \quad (1)$$

In this study, we focus on shape uncertainty given a model $\theta_j$, i.e., to learn the distribution $\mathcal{P}(s|\theta_j, x)$. The uncertainty in Eq.(1) can be derived by training multiple models.

### B. CQ-AE

We first introduce the CQ-AE which can be considered a simplification of CQ-VAE. Fig. 1 shows the framework of CQ-AE applied on the Fashion-MNIST dataset. The encoder consists of a series of convolution layers, and the decoder has a series of transposed convolution layers.

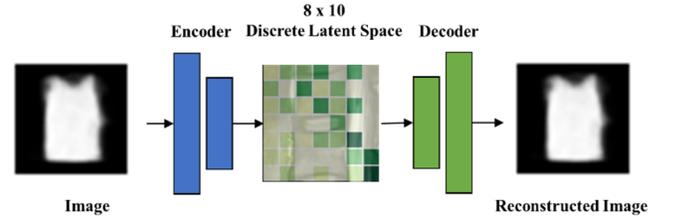

Fig. 1. The architecture of CQ-AE

The discrete latent space is represented by a M×N matrix denoted by $z$. For Fashion-MNIST dataset, we set M=8 and N=10. Each row of the matrix $z$ is a softmax output from the encoder, i.e. $\sum_m z[m,:] = 1$ and $0 \leq z[m,n] \leq 1$ where $z[m,n]$ is the element in the m-th row and the n-th column of $z$. Between the matrix $z$ and the decoder, there is a special operation, given by:

$$z' = z \times c \quad (2)$$

where $c$ is a pre-defined constant vector of N elements and $z'$ is a vector living in a continuous M-dimensional latent space. This special operation is the bridge from the discrete latent space to the continuous latent space.

The loss of the CQ-AE is given by:

$$L_{CQAE} = \|\hat{x} - x\|^2 + \alpha \cdot H(z) \quad (3)$$

where $H(z) = -\sum_{m,n} z[m,n]\log(z[m,n])$ is entropy of $z$, $\alpha$ is a weight scalar, and $\|\hat{x} - x\|^2$ is reconstruction error. By using a large positive scalar $\alpha$ and minimizing the entropy $H(z)$, each row of $z$ will become almost one-hot: one of the elements is close to 1 and the others are close to zeros.

Once $z$ becomes one-hot, there only exists a finite number of $z'$ vectors in the continuous M-dimensional latent space. Actually, the number of unique $z'$ vectors is $N^M$.

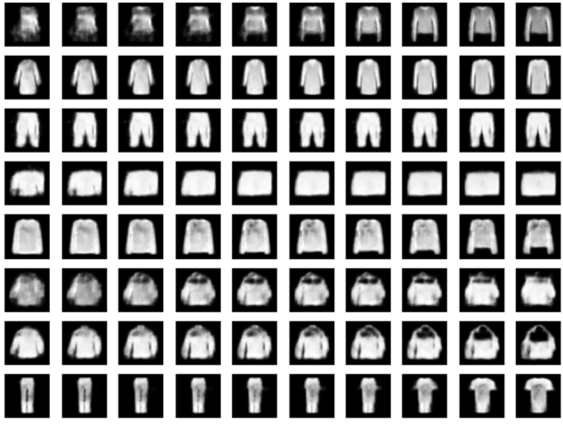

Fig. 2. Images generated by random sampling

From another perspective, if we quantize each dimension of the continuous M-dimensional (latent) space by $N$ discrete coordinates, then we create a discrete space (e.g. a grid) that contains $N^M$ unique points. Each point in the discrete space is represented by a binary matrix $z$, and it can be mapped back to the continuous space by the operation $z' = z \times c$. Thus, the CQ-AE has a discrete latent space created from a continuous space by coordinate quantization.

Once the CQ-AE is trained, it can be used to generate new images from random binary matrices $z$ (with 0s and 1s inside), as shown in Fig. 2. The Fashion-MNIST dataset has 60,000 samples in the training set. The CQ-AE can represent $N^M = 10^8 \gg 60000$ unique vectors/objects in the latent space, and therefore, it can generate $N^M$ different images, most of which do not exist in the training set.

### C. CQ-VAE

The framework of CQ-VAE is shown in Fig. 3, which is an upgrade of CQ-AE, for uncertain estimation. It has a shape space $\mathbb{S}$ and a discrete latent space $\mathbb{Z}^{M \times N}$. Any $M \times N$ binary matrix $z$ is in the space $\mathbb{Z}^{M \times N}$, and it can be mapped into a vector $z'$ in a continues latent space $\mathbb{R}^M$ by $z' = z \times c$.

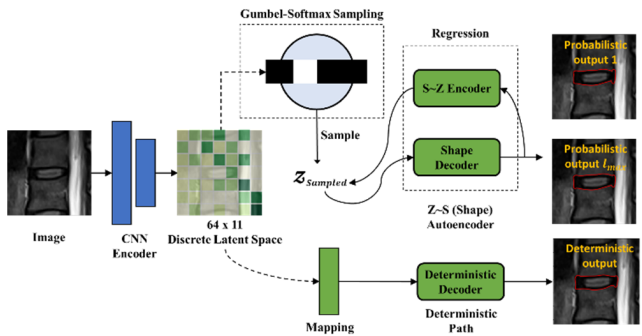

Fig. 3. The architecture of CQ-VAE

For this lumbar disk application, a shape $s \in \mathbb{S}$ is a 2D contour with a pre-defined number of points. For other applications, a shape could be a mesh with a pre-defined number of nodes and elements. Shape correspondence between patients means that: the i-th point of the shape $s_1$ from patient-1 corresponds to the i-th point of the shape $s_2$ from patient-2.

In the following paragraphs, we will introduce the objective of CQ-VAE, starting from a vanilla VAE.

**VAE**. A VAE consists of a discriminative network $q(z|x)$ (a.k.a. encoder) and an inference network $p(s|z)$, where $s$ is a shape, $s \in \mathbb{S}$, and $z$ is a latent code, $z \in \mathbb{Z}$. The idealized objective of a VAE is the marginalized log-likelihood $\log p(x)$, which can be approximated by the evidence lower bound (ELBO) [6]:

$$\log p(x) \geq \mathbb{E}_{(z,s) \sim q(z,s|x)}[\log p(x|z) + \log p(s|z)]$$
$$- D_{KL}(q(z|x) \| p(z)) \quad (4)$$

In our application, we are not interested in reconstructing the image $x$, then $\log p(x|z)$ is removed. Thus, VAE objective is:

$$U_{VAE} = \mathbb{E}_{(z,s) \sim q(z,s|x)}[\log p(s|z)]$$
$$- D_{KL}(q(z|x) \| p(z)) \quad (5)$$

where $p(z)$ is the prior distribution, e.g., a uniform discrete distribution in our model. $q(z|x)$ represents a probabilistic encoder to generate $z$, given $x$. $p(s|z)$ represents a deterministic decoder to generate $s$, given $z$. By definition, we have $q(z,s|x) \triangleq q(z|x)p(s|z)$. Thus, the first term, $\mathbb{E}_{(z,s) \sim q(z,s|x)}[\log p(s|z)]$, is used to maximize the log-likelihood of the generated shapes, while the remaining one is the KL divergence between the latent space distribution and a fixed prior uniform distribution.

**Autoencoder (AE)**. Shape space $\mathbb{S}$ is discrete because the latent space $\mathbb{Z}$ is discrete and each shape $s$ is generated from a latent code $z$. Thus, we employ an autoencoder $q(z|s)$ to enforce one-to-one correspondence between $\mathbb{Z}$ and $\mathbb{S}$, i.e., each shape $s$ has a unique latent representation $z$, and each $z$ is decoded only to one unique shape $s$. The auto-encoding process is expressed as:

$$z \rightarrow s \rightarrow \hat{z} \cong z \quad (6)$$

The AE objective is:

$$U_{AE} = \mathbb{E}_{z \sim q(z|x)}\left[\mathbb{E}_{s \sim p(s|z)}[\log q(z|s)]\right]$$
$$= \mathbb{E}_{(z,s) \sim q(z,s|x)}[\log q(z|s)] \quad (7)$$

**Probabilistic Path for Shape Regression**. In our application, we have shape data from human experts, and therefore we need to incorporate this information into the objective. Let $\tilde{s}$ denote a shape from a human expert, and let $\tilde{p}(\tilde{s}|x)$ denote the shape distribution from human experts (see section E for details). Then, the regression objective is:

$$U_{reg} = \mathbb{E}_{\tilde{s} \sim \tilde{p}(\tilde{s}|x)}[\log q(\tilde{s}|x)] \quad (8)$$

where $q(s|x)$ is the probabilistic path in CQ-VAE from input $x$ to output shape $s = \tilde{s}$. $q(s|x) = q(z,s|x)$ because of one-to-one correspondence between shape and laten code.

**Deterministic Path for Shape Regression**. We add a deterministic path to the model to output the best shape estimation $s_{best}$. This output shape is intended to be compared with the "best" ground-truth shape $s^*$ which is the consensus among human experts. The objective is

$$U_{best} = \log q(s_{best}|x) \qquad (9)$$

where $q(s_{best}|x)$ represents the path from $x$ to $s_{best}$.

**The total objective** of our CQ-VAE model is:

$$U_{CQ\_VAE} = U_{VAE} + \alpha U_{AE} + U_{reg} + \beta U_{best} \qquad (10)$$

where coefficients $\alpha$ and $\beta$ will be determined on validation set. The terms $U_{VAE} + \alpha U_{AE}$ are derived from unsupervised learning given the images, which describe the generative process. The terms $U_{reg} + \beta U_{best}$ are derived from supervised learning given the images and shapes. There is no explicit coefficient before $U_{reg}$, and we use a shape sampling-matching method (section E) to implicitly control its weight.

*D. Gumbel-softmax sampling*

The advantage of discrete representation is very obvious because any continuous probability distribution can be discretized with a sufficient resolution, and therefore, we do not need an analytical form of the distribution function. However, it also causes a problem: the gradient of the objective cannot directly backpropagate through the discrete representation. Fortunately, it can be resolved by using Gumbel-softmax sampling. The Gumbel-softmax distribution is a continuous distribution that can be used to draw samples from a discrete distribution [10] by:

$$z = one\_hot(\underset{i}{\mathrm{argmax}}[g_i + \log \pi_i]) \qquad (11)$$

where $g_1, \ldots, g_N$ are i.i.d. samples drawn from Gumbel(0,1) and $\pi_i$ is the i-th element in a row of the $M \times N$ probability matrix. By replacing one-hot with softmax in Eq.(11), the sampling process becomes differentiable with respect to $\pi_i$.

*E. Shape Sampling and Shape Matching*

For each disk image $x$, we have three shapes $s^{(1)}, s^{(2)}, s^{(3)}$ from three human experts, and one best shape $s^*$ from the consensus of the experts. From the three shapes, we can derive a distribution of shapes, by sampling shapes from the linear combination $\tilde{s} = \sum_{i=1}^{3} \alpha_i s^{(i)}$ where $0 \leq \alpha_i \leq 1$, $\sum_{i=1}^{3} \alpha_i = 1$ and $\alpha_i$ is a random number from a uniform distribution. Thus, the "ground-truth" shape distribution, $\tilde{p}(\tilde{s}|x)$, is represented by a large number of sampled shapes $\{\tilde{s}^{(1)} \ldots \tilde{s}^{(k)} \ldots \tilde{s}^{(k_{max})}\}$. It is impractical to have $k_{max}$ shapes from $k_{max}$ human experts, and therefore, this shape sampling approach is the best we can do in the real situation.

For each disk image $x$, the CQ-VAE has a discrete probability distribution $q(s|x) = q(z,s|x) \triangleq q(z|x)p(s|z)$. In theory, the generative model $q(s|x)$ should be able to reproduce the shapes in the training set and also generate new shapes that are not seen in the training set. A large number of shapes $\{s^{(1)} \ldots s^{(l)} \ldots s^{(l_{max})}\}$ can be sampled from the distribution $q(s|x)$. If $l_{max}$ is large enough ($l_{max} > k_{max}$), then $\{\tilde{s}^{(1)} \ldots \tilde{s}^{(k_{max})}\}$ should be a subset of $\{s^{(1)} \ldots s^{(l_{max})}\}$. For training, we need the correspondence $s^{(l(k))} \leftrightarrow \tilde{s}^{(k)}$ in order to compute the objective in Eq.(8), and we apply global greedy assignment algorithm [11] to find the best shape matching $l(k)$, and the Euclidean distance $\|s^{(l(k))} - \tilde{s}^{(k)}\|$ is used in the algorithm for shape matching.

The shape-matching process is illustrated in Fig. 4. A pure random matching algorithm will result in a model learning only the mean shape of the training set. Instead, the shape matching algorithm will pair each "ground truth" shape with the best-matching shape from CQ-VAE. By setting $l_{max} > k_{max}$, the model allocates $l_{max} - k_{max}$ shapes to represent new shapes not in training set and $k_{max}$ shapes to be close to the training shapes. In this way, the model can make a trade-off between the ability to generate new shapes and the ability to represent training shapes, which sets an implicit weight for $U_{reg}$ in the total objective.

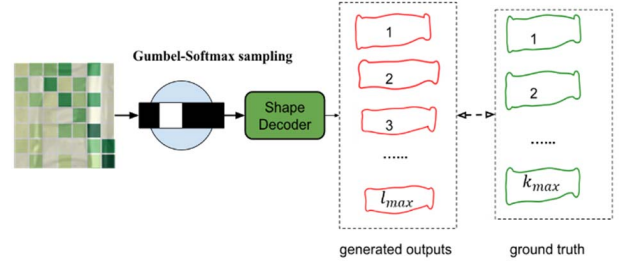

Fig. 4. Shape-matching

### III. EXPERIMENT

We conducted qualitative and quantitative evaluations, including uncertainty estimation, uncertainty visualization, generative capability test, to demonstrate our model.

*A. Dataset*

The dataset consists of de-identified MRI lumbar spine images of 100 patients from our medical school. Three human experts manually annotated the boundaries and landmarks of the lumbar disks and vertebrae on the mid-sagittal images, by following the protocol in [5]. The best annotation for each image was obtained through discussion to reach a consensus among the experts. To build shape correspondence between patients, each boundary/contour of a disk/vertebrae is divided into segments between landmarks, and each segment is resampled into a pre-defined number of points. As a result, each disk shape contains 176 points. Data from 80 patients are used for training; and data from the remaining patients are used for testing. For the train-test data splitting, we did random selection and also made sure that the patient age distributions in the two sets are similar. Each image contains 5 disks. For our experiments, we cropped each image into squared regions, and each region contains a disk at the center. Then, each cropped region was resized to 128 ×128 pixels.

**Data Augmentation on training set.** We applied PCA to the lumbar disk shapes in the training set to build a statistical shape model (SSM), which has a mean shape and modes of shape variations covering about 80% of the total variations. We used this SSM to generate virtual disk shapes, similar to the work in [12]. Given an image $x$ of a disk, the consensus shape $s^*$ of the disk from the experts, and a virtual shape $\tilde{s}$ randomly sampled from the SSM ($s^*$ and $\tilde{s}$ can be very different), an image $\tilde{x}$ of this virtual shape $\tilde{s}$ is generated by applying a thin-plate-spline (TPS) transform to the image $x$. The parameters of the TPS are determined by the transform between $s^*$ and $\tilde{s}$. The shapes from individual experts are also transformed to the space of the new image $\tilde{x}$ by the TPS transform. In total, 1200 disk images with shapes are obtained after data augmentation on the training set. Data augmentation is not used on test set.

### B. Uncertainty Measured by Shape-Variation and Entropy

The direct measurement of uncertainty is the probability distribution of shapes given an input image. However, the "ground-truth" shape distribution cannot be used for the direct comparison because $\tilde{p}(\tilde{s}|x)$ is represented by a large number of sampled shapes $\{\tilde{s}^{(1)} ... \tilde{s}^{(k)} ... \tilde{s}^{(k_{max})}\}$. Thus, we use two indirect measurements: shape-variation and entropy.

Shape-variation is calculated as follows. Let $\tilde{s}^{(avg)}$ denote the mean shape of the sampled shapes, and $\tilde{s}[j]$ denote the j-th point of shape $\tilde{s}$, then the "ground-truth" shape-variation $var(\tilde{s})$ is defined by

$$var(\tilde{s}) = \frac{1}{j_{max}*k_{max}}\sum_j \sum_k \left\|\tilde{s}^{(k)}[j] - \tilde{s}^{(avg)}[j]\right\| \quad (12)$$

where $j$ is from 1 to $j_{max}$ and $k$ is from 1 to $k_{max}$. $var(\tilde{s})$ is zero if all of the sampled shapes are identical, which means zero uncertainty/variation among the human experts-reconstructed shapes. Similar to Eq.(12), we can also calculate $var(s)$ using the shapes $\{s^{(1)} ... s^{(l)} ... s^{(l_{max})}\}$ sampled from $q(s|x)$ in the model, given an input image. We name $var(s)$ model-computed shape-variation, which is computed by:

$$var(s) = \frac{1}{j_{max}*l_{max}}\sum_j \sum_l \left\|s^{(l)}[j] - s^{(avg)}[j]\right\| \quad (13)$$

We compared $var(s)$ and $var(\tilde{s})$ for every image in the test set. As shown in Fig. 5, the ground-truth and model-computed shape-variation are positively correlated and the correlation coefficient is 0.423.

The second uncertainty measurement is the entropy of $q(z|x)$ which is explicitly represented by a $M \times N$ probability matrix. Intuitively, if $q(z|x)$ is a binary matrix, then its entropy is at its minimum and only a single shape can be generated, which means uncertainty is very low. If $q(z|x)$ is close to a uniform distribution, then entropy is at its maximum and a large number of different shapes can be generated, which means uncertainty is very high. The entropy for every image in the test set is computed, as shown in Fig. 6.

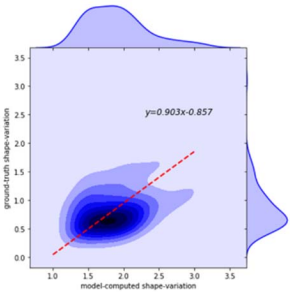
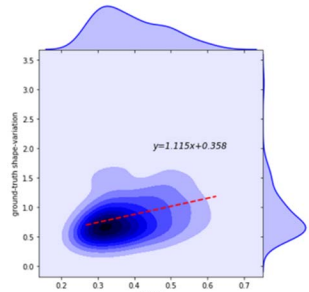

Fig. 5. Model-computed vs "Ground-truth" shape-variations

Fig. 6. Model-computed entropy vs "Ground-truth" shape-variation

Both uncertainty measurements are in good agreement with the "ground-truth" shape-variation on the test set. The results quantitatively demonstrate that CQ-VAE can indeed learn shape uncertainty in images. From Fig. 5, it is observed that the range of $var(s)$ is larger than the range of $var(\tilde{s})$ because $var(s)$ is related to not only data uncertainty but also the uncertainty of model output (see Eq.(1) and section III-D).

### C. Visualization of Shape-Variation

In this section, we will visualize shape-variation for four representative disk images in order to present the uncertainty in an intuitive way. For the visualization of the shape-variation, we draw 100 samples from CQ-VAE to obtain 100 possible shapes for a disk image. The visualization would be really confusing if these 100 shapes are plotted all together over the image. Instead, the model-computed average shape $s^{(avg)}$ is plotted with model-computed variation on each point. Similarity, the "ground-truth" average shape $\tilde{s}^{(avg)}$ is plotted with "ground-truth" variation on each point.

The results of four representative disks are visualized in Fig. 7. The left column shows the "ground-truth" heatmaps. The middle column shows the model-computed heatmaps. Red color highlights low-variation (low-uncertainty) regions where points of the shapes are more likely to be clustered; blue color indicates high-variation (high-uncertainty) regions. We can see that the overall shape-variation along the disk boundary is similar between "ground-truth" and model-computed. Also, the points (i.e. local shape) near the boundary between disk and vertebra have relatively low-variation because the boundary is relatively straight and easy to identify; the points near the anterior and posterior of the disk have high-variation because the local shape is highly curved and there is large disagreement between human experts.

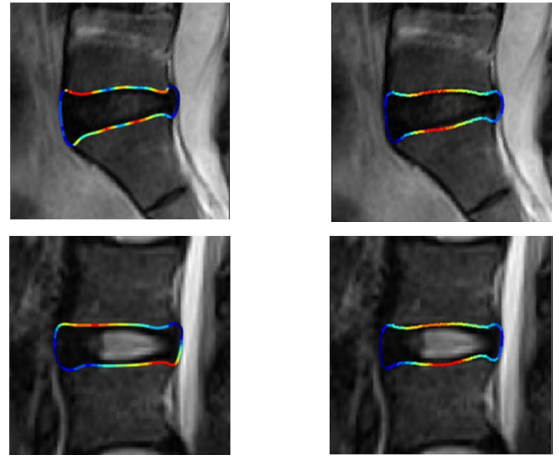

Fig. 7. Visualization of shape-variation for four representative disks. The first column shows the "ground-truth" results. The second column shows the model-computed results.

We note that that $q(s|x) = \mathcal{P}(s|\theta_j, x)$ and $\tilde{p}(\tilde{s}|x) \approx \mathcal{P}(s|x)$ in Eq.(1). Thus, the learned distribution $q(s|x)$ in CQ-VAE is not exactly the same as $\tilde{p}(\tilde{s}|x)$ on the test set.

### D. Uncertainty v.s. Bias

By definition, uncertainty is different from bias which is the discrepancy between the best estimation $s_{best}$ from CQ-VAE and the consensus shape $s^*$ from the human experts. It would be interesting to study the relationship between uncertainty and bias. The bias of a shape $s$ from CQ-VAE is:

$$bias(s) = \frac{1}{j_{max}} \sum_j \|s[j] - s^*[j]\| \qquad (14)$$

We computed $bias(s_{best})$ and compared it with the two uncertainty measures: model-computed shape-variation and entropy. The results on the test set are shown in Fig. 8 and 9. The trend is obvious: as uncertainty becomes higher, the bias increases. The correlation coefficient between bias and model-computed shape-variation is 0.806. The correlation coefficient between bias and entropy is 0.545. The results demonstrate that our CQ-VAE model knows the uncertainty of its output.

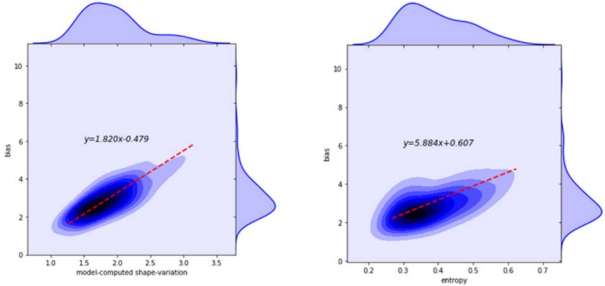

Fig. 8. bias vs model-computed shape-variation

Fig. 9. bias vs model-computed entropy

*E. Generating Shapes*

CQ-VAE is a generative model, and theoretically, it can represent $N^M$ unique samples in the $M \times N$ discrete latent space. In our application, M=64 and N=11 which enables CQ-VAE to represent and generate $11^{64}$ shapes, which are more than enough for the application. Examples of the generated shapes are shown in Fig. 10.

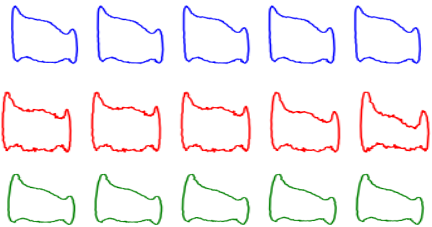

Fig.10. Shapes generated by random sampling

## IV. CONCLUSION

In this work, we developed CQ-VAE, a novel Coordinate-Quantized Variational Autoencoder to learn a representation of uncertainty, and it was applied for the application of lumbar disk shape analysis from MR images. The discrete representation enables CQ-VAE to represent any discrete probability distribution by coordinate-quantization. The difficulty of backpropagating through the discrete representation is resolved by using Gumbel-softmax sampling. A trade-off between the ability to generate new shapes and the ability to represent training shapes, is made by using a shape sampling and matching strategy. The results of the application quantitatively demonstrate that CQ-VAE can learn shape uncertainty in images. CQ-VAE is a general framework for uncertainty representation learning, and it may be applied for other applications for uncertainty estimation as long as the probability distribution of the target is discretized.

Uncertainty estimation is very challenging compared to common supervised learning tasks, and it is difficult to directly use "ground-truth" uncertainty scores in the objective/loss function for uncertainty score regression. For example, it would be ideal to directly learn the relationship between model-computed shape-variation and bias by using a loss term measuring the error of linear regression between shape-variation and bias. However, in many applications, a large model with many training epochs is usually used, and therefore the bias could approach zero on the training set, which makes the linear regression to be useless. Thus, our future work will include the investigation of directly using "ground-truth" uncertainty scores in the objective and refining the evaluation approach. Nevertheless, to our knowledge, our work is the first one that builds a novel VAE for estimating object shape uncertainty in medical images.

ACKNOWLEDGMENT

This research was supported in part by Amazon AWS Machine Learning Research Award.